\documentclass{article} 
\usepackage{compositional_metalearning,times}
\usepackage[pdftex]{graphicx}

\usepackage{amsmath,amsfonts,bm}

\def\eqref#1{equation~\ref{#1}}

\def\1{\bm{1}}

\def\rk{{\textnormal{k}}}

\def\ru{{\textnormal{u}}}

\def\rz{{\textnormal{z}}}

\def\rvm{{\mathbf{m}}}

\def\rvx{{\mathbf{x}}}
\def\rvy{{\mathbf{y}}}
\def\rvz{{\mathbf{z}}}

\def\vmu{{\bm{\mu}}}
\def\vtheta{{\bm{\theta}}}
\def\vphi{{\bm{\phi}}}

\def\ve{{\bm{e}}}

\def\vg{{\bm{g}}}

\def\vm{{\bm{m}}}

\def\mI{{\bm{I}}}

\def\mW{{\bm{W}}}

\DeclareMathAlphabet{\mathsfit}{\encodingdefault}{\sfdefault}{m}{sl}
\SetMathAlphabet{\mathsfit}{bold}{\encodingdefault}{\sfdefault}{bx}{n}

\newcommand{\R}{\mathbb{R}}

\usepackage{hyperref}
\usepackage{url}

\title{Compositional meta-learning \\ through probabilistic task inference}

\author{
{\bf Jacob J.~W.~Bakermans\textsuperscript{1,$\dagger$}, 
Pablo Tano\textsuperscript{1,2}, 
Reidar Riveland\textsuperscript{1,3},}\\
{\bf Charles Findling\textsuperscript{1,*}, 
Alexandre Pouget\textsuperscript{1,*}} 
 \\[0.75em]
\textsuperscript{1}Department of Basic Neurosciences, University of Geneva, Switzerland \\
\textsuperscript{2}Champalimaud Research, Champalimaud Foundation, Portugal \\
\textsuperscript{3}Sainsbury Wellcome Centre, University College London, United Kingdom \\[0.75em]
\textsuperscript{*} These authors contributed equally \\
\textsuperscript{$\dagger$} Correspondence: \texttt{jacob.bakermans@gmail.com}
}

\iclrfinalcopy 
\begin{document}

\maketitle

\begin{abstract}
To solve a new task from minimal experience, it is essential to effectively reuse knowledge from previous tasks, a problem known as meta-learning. Compositional solutions, where common elements of computation are flexibly recombined into new configurations, are particularly well-suited for meta-learning. Here, we propose a compositional meta-learning model that explicitly represents tasks as structured combinations of reusable computations. We achieve this by learning a generative model that captures the underlying components and their statistics shared across a family of tasks. This approach transforms learning a new task into a probabilistic inference problem, which allows for finding solutions without parameter updates through highly constrained hypothesis testing. Our model successfully recovers ground truth components and statistics in rule learning and motor learning tasks. We then demonstrate its ability to quickly infer new solutions from just single examples. Together, our framework joins the expressivity of neural networks with the data-efficiency of probabilistic inference to achieve rapid compositional meta-learning.
\end{abstract}

\section{Introduction}

Learning a new task rarely requires starting from scratch. For example, to cook a new dish you can rely on previously learned techniques like chopping or frying, or to play a new piano sonata you can incorporate fragments of familiar movements. Having learned a set of related “training” tasks (or dishes or sonatas) facilitates the acquisition of the new “test” task, a process known as “learning-to-learn” or “meta-learning” \citep{harlowFormationLearningSets1949, hospedalesMetaLearningNeuralNetworks2020}. But these examples highlight a particularly powerful way of meta-learning: both involve reusing components from previous tasks in new configurations \citep{lakeHumanlevelConceptLearning2015, lakeBuildingMachinesThat2017}. This is a form of compositionality, which affords combinatorial generalisation by reusing a small number of elements in endless new combinations \citep{fodorLanguageThought1975, battagliaRelationalInductiveBiases2018}. Here, we will propose a model that exploits compositionality to discover common components across solutions and learn how to combine those components to rapidly solve new tasks.

Such reusable components of computation emerge spontaneously in recurrent neural network (RNN) models trained to perform many tasks, a setting called “multi-task learning” \citep{caruanaMultitaskLearning1997, yuMetaWorldBenchmarkEvaluation2020}. For example, in an RNN trained on a suite of rule learning tasks, the same subnetwork activates across tasks that require evidence integration, and another subnetwork when producing a response \citep{yangTaskRepresentationsNeural2019}. These components are implemented as “dynamical motifs”, recurrent dynamics that carry out specific computations and are shared across tasks that require those computations \citep{driscollFlexibleMultitaskComputation2024}. Importantly, these motifs support meta-learning: a new task that relies on a previously learned motif is acquired much faster than a task that requires a new motif. Our model builds on such dynamical motifs through a set of learnable modules that each implement an isolated but reusable computation.

To solve the task at hand, our model needs to combine these modules. But these combinations are often richly structured \citep{suttonMDPsSemiMDPsFramework1999a, markowitzSpontaneousBehaviourStructured2023}. For example, the evidence integration module may usually be followed by the response module, but never the other way around. We therefore introduce a separate learnable model, which we call the gating network, that selects which module is activated when. This model acts like the gating function in a mixture-of-experts setup \citep{jacobsAdaptiveMixturesLocal1991}, or like the routing function in modular networks \citep{rosenbaumRoutingNetworksAdaptive2017}. Importantly, this separation between modules and gating provides a strong inductive bias for the gating network to focus on the statistics of module combinations, rather than on the specifics of module dynamics. This architecture therefore encourages the gating network to learn the grammar that generates tasks, while the modules learn the tasks’ syllables.

Together, the gating network and the modules thus learn a generative model of the “train” tasks. By casting our framework explicitly as a probabilistic generative model, we can solve new “test” tasks through probabilistic inference. That is fundamentally different from common meta-learning approaches \citep{finnModelAgnosticMetaLearningFast2017, liLearningGeneralizeMetaLearning2018, nicholFirstOrderMetaLearningAlgorithms2018, rosenbaumRoutingNetworksAdaptive2017, pontiCombiningModularSkills2022, chitnisLearningQuicklyPlan2019} which, through architecture or training, aim to minimise the parameter updates required to learn test tasks. By inferring rather than learning new solutions, we avoid parameter updates altogether. Instead, we compose solutions from modules, constrained by their statistics, guided by task feedback.

We show that we can simultaneously learn the gating network and modules by recovering ground truth modules and statistics in two abstract domains. Then, we demonstrate an inference procedure on our learned generative model that can infer new tasks from minimal experience, even when task feedback is sparse. Together, these results provide a framework for rapid acquisition of new tasks through compositional meta-learning.

\section{Results}
\subsection{A generative model of tasks} \label{sec:model}
Our goal is to rapidly find a solution to a new (or test) task, after having encountered a set of previous (or training) tasks. We define a task as a data-generating process that yields episodes of input-output pairs $\{\rvx_t,\rvy_t\}_{t=1...T}$. Our model needs to discover the underlying process to produce the correct output $\rvy_t$ given input $\rvx_t$ for the test task, from a minimal number of episodes. To achieve that, it needs to learn the commonalities across the training tasks, and figure out how to apply those in the test task. Importantly, we assume that many real-world tasks are modular: they generate data through varying combinations of sub-processes or modules. Our model therefore needs to learn two key characteristics of the training tasks. First, it must isolate the modules to learn within-module dynamics; and second, it must extract how they are combined to learn between-module dynamics.

\begin{figure}[h]
\begin{center}
\includegraphics[scale=0.75]{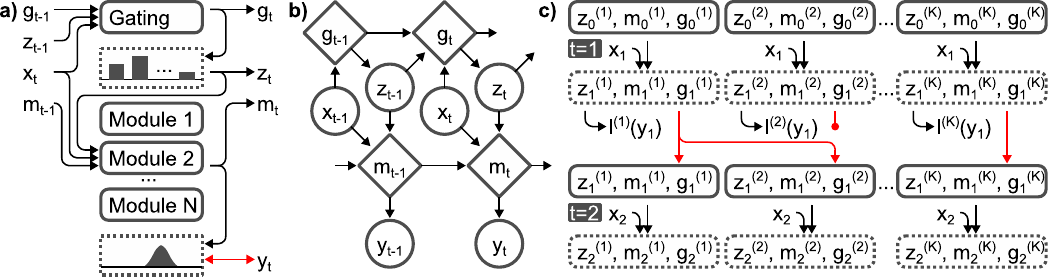}
\end{center}
\caption{Model overview. \textbf{a)} Model architecture. The model consists of a gating RNN that for a given gating hidden state $\vg_{t-1}$, previously activated module $\rz_{t-1}$, and input $\rvx_t$, parameterises a discrete probability distribution from which the currently active module RNN $\rz_t$ is sampled. The selected module RNN processes the input $\rvx_t$ and module hidden state $\rvm_{t-1}$ to define an output distribution for $\rvy_t$. \textbf{b)} Graphical model. The model learns a probabilistic generative process with stochastic variables (circles) $\rz_t$ and $\rvy_t$ that depend on input $\rvx_t$ and the deterministic model hidden states (diamonds) $\vg_t$ and $\vm_t$. Conceptually, this expands a HMM by replacing the transition and emission matrices by input-dependent RNNs. \textbf{c)} Particle filter schematic. To perform inference in this generative model, we define a particle system of $K$ particles (top row). At a given timestep, we sample module activations to calculate likelihoods $l^{(i)}$ of data $\rvy_t$ for each particle (second row). We resample particles from these (normalised) likelihoods (red arrows: particle $(1)$ is sampled twice, whereas particle $(2)$ is terminated) to reflect the module posterior $p(\rz_t|\rvy_{1:t})$ (third row), and continue this process for the next timestep (bottom row).}
\label{fig:model}
\end{figure}

Our model separates these two learning objectives through its architecture (Figure \ref{fig:model}a; Appendix \ref{app:model}). The between-module dynamics are captured by a gating network, which selects from a set of module networks that each capture different within-module dynamics. We implement this through a gating RNN $G_\vtheta$ that decides at each timestep which module $\rz_t$ is activated from a set of module RNNs $\{M_\vphi^z\}_{z=1...N}$
\begin{equation} \label{eq:1}
\vg_t = G_\vtheta(\rvx_t,\vg_{t-1},\rz_{t-1})
\end{equation}
\begin{equation} \label{eq:2}
\rz_t \sim \text{Cat}(\mW_G \vg_t)
\end{equation}
where $\mW_G$ is a linear projection from the hidden state $\vg_t$ of $G_\vtheta$ to the number of modules $N$ and $\text{Cat}$ denotes a categorical distribution. The selected module gets to process the input to produce an output
\begin{equation} \label{eq:3}
\vm_t = M_\vphi^{\rz_t}(\rvx_t,\vm_{t-1})
\end{equation}
\begin{equation} \label{eq:4}
\rvy_t \sim \text{MVN}(\mW_M \vm_t, \sigma \mI)
\end{equation}
where $\mW_M$ is a linear projection from the hidden state $\vm_t$ of $M_\vphi^z$ to the output dimension $d_y$, $\sigma$ is a learnable standard deviation with $\mI$ the $d_y$ dimensional identity matrix, and $\text{MVN}$ denotes a multivariate normal distribution. For notational convenience, we’ll group all learnable parameters as $\Lambda=\{\sigma, \vtheta, \vphi, \mW_G, \mW_M\}$.

Crucially, we don’t provide any of the networks with the task identity. Whereas indicating the current task is standard in brain-inspired multi-task learning \citep{yangTaskRepresentationsNeural2019, dunckerOrganizingRecurrentNetwork2020, martonEfficientRobustMultitask2022, rivelandNaturalLanguageInstructions2024, driscollFlexibleMultitaskComputation2024}, our model learns from task output feedback alone. The resulting model therefore doesn’t specify how to solve one particular task. Instead, it learns the underlying dynamics across all tasks, formalised as a probabilistic generative model in Equations \ref{eq:1}-\ref{eq:4} (Figure \ref{fig:model}b). Once that generative model has been learned from the training tasks, finding a solution to the test task is a matter of inference rather than learning - there’s no need for parameter updates. Concretely, having learned the within-module dynamics, solving the test task only requires finding the module sequence that best explains the test task data. And having learned the between-module dynamics, the space of possible sequences will be highly constrained. This process resembles learning and inference in a classic hidden markov model (HMM), but one where the transition matrix is replaced by the gating RNN and the emission matrix by the module RNNs. These RNNs greatly enhance expressivity: the gating RNN can learn long-distance non-Markovian dependencies, and the module RNNs can learn arbitrary emission functions. Meanwhile, we can still rely on the efficient inference machinery afforded by probabilistic models. 

We apply these probabilistic inference methods both for inferring the solution for the test task and to compute the loss function used to train the RNNs. Specifically, we use particle filtering \citep{gordonNovelApproachNonlinear1993, doucetTutorialParticleFiltering2009} to obtain the posterior module selection of a test episode and to compute the marginal likelihood of the training episodes (although our model is agnostic to the choice of approximate inference method - others may also work). Briefly (Figure \ref{fig:model}c; Appendix \ref{app:inference}), given a particle system $\{k_{t-1}\}_{k=1...K}$ where 
$k_{t-1}=\{\rz_{1:t-1}^{(k)},\vg_{1:t-1}^{(k)},\vm_{1:t-1}^{(k)}\}$ we sample a module transition through Equations \ref{eq:1}-\ref{eq:2} to get module selection $\{\rz_t^{(k)}\}_{k=1...K}$. By evaluating the sampled modules via Equations \ref{eq:3}-\ref{eq:4}, we obtain output mean $\{\vmu_t^{(k)}\}_{k=1...K}$ where $\vmu_t=\mW_M \vm_t$, which provides the particle’s likelihood of the current target output $\rvy_t$:
\begin{equation} \label{eq:5}
l_t^{(k)}=p(\rvy_t|\rz_t^{(k)}; \Lambda) = p(\rvy_t|\vmu_t^{(k)};\Lambda) 
\end{equation}
We then sample particles for propagation to the next timestep from the (normalised) likelihoods 
\begin{equation} \label{eq:6}
\rk_t \sim \text{Cat}(l_t^{(k)} / \sum_{i=1}^K l_t^{(i)})
\end{equation}
via stratified resampling (Appendix \ref{app:inference}). This particle system provides us with two important quantities. The distribution of modules selected by resampled particles from Equation \ref{eq:6} approximates the posterior $p(\rz_t|\rvy_{1:t})$. We need that to infer the best sequence of modules for the test task. The particle likelihood before resampling in Equations \ref{eq:5} determines the loss for learning the model parameters on the training tasks. We need that to calculate the marginal likelihood at each timestep 
\begin{equation} \label{eq:7}
p(\rvy_t|\rvy_{1:t-1};\Lambda)=\frac{1}{K}\sum_{i=1}^K l_t^{(i)}
\end{equation}
so that we can get the marginal likelihood across the whole timeseries 
\begin{equation} \label{eq:8}
L=p(\rvy_{1:T};\Lambda)=\prod_{t=1}^T \frac{1}{K} \sum_{i=1}^K l_t^{(i)}
\end{equation}
as the training loss. We optimise model parameters $\Lambda=\{\sigma, \vtheta, \vphi, \mW_G, \mW_M\}$ through gradient descent on $-\text{log}(L)$, backpropagating the loss through the particle filter on the training tasks (Appendix \ref{app:inference}). We use the gumbel-softmax reparameterisation trick to calculate gradients through the sample of Equation \ref{eq:2}.

In summary, we learn a probabilistic generative model that separates within-module dynamics (acting as ‘task syllables’) from between-module dynamics (to form a ‘task grammar’). We train this model by maximising the marginal likelihood of the training tasks through backpropagation. Then we find the best module sequence to solve the test task by probabilistic inference. Importantly, this means that test tasks are solved without any parameter updates. In the remainder of this paper, we will demonstrate the power of this approach on rule learning and motor learning tasks.

\subsection{Recovery in abstract rule learning} \label{sec:rule-learning}
We first apply our model to an abstract rule learning task to show it can recover ground truth within-module and across-module dynamics. This task mimics the hardest variants of context-based rule switching because the model needs to simultaneously learn what the rules are and when to apply them, from uninformative input. The data-generating process is given by
\begin{equation} \label{eq:9}
\rvy_t=S_s(\rvy_{t-1})+\rvx_t
\end{equation}
where $\rvy_t, \rvx_t \in \R^6$, $\rvx_t \sim \mathcal{N}(0,\mI_6)$, $\rvy_0=0$, and $\{S_s\}_{s=0...5}$ is the set of 6D shift operations that shift the value of vector entry $i$ to entry $\text{mod}(i+s,6)$. A task is defined as a sequence of shift operations that determines the input-output mapping at each timestep. Importantly, these sequences have a very specific structure. Each task consists of three shift operations, and each shift operation repeats a fixed number of timesteps: 3 timesteps for $S_0$ and $S_1$, 4 timesteps for $S_2$ and $S_3$, and 5 timesteps for $S_4$ and $S_5$. Given this structure, an example task may sequence $S_1$ then $S_4$ then $S_2$ so that $s$ takes on the values 1,1,1,4,4,4,4,4,2,2,2,2 over timesteps. Thus, from episodes generated by the training tasks, the module RNNs must each learn one of the shift operations, and the gating RNN needs to learn the sequence regularities.

\begin{figure}[h]
\begin{center}
\includegraphics[scale=0.75]{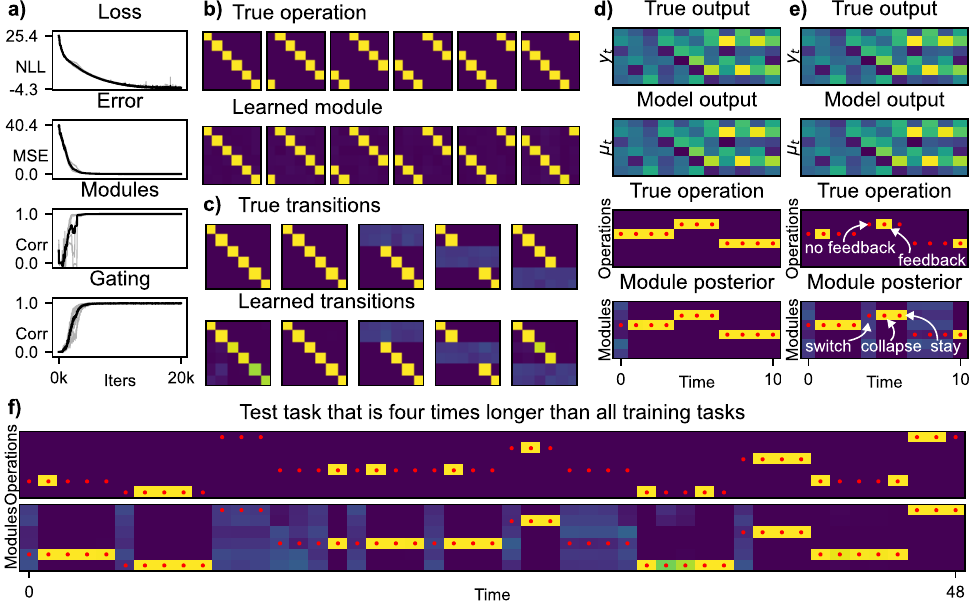}
\end{center}
\caption{Rule learning. \textbf{a)} Learning curves. The training loss (negative log marginal likelihood) and task performance (mean squared error) decrease while the module and gating accuracy (correlation with ground truth operations and transitions) plateaus at 1 (grey lines: five individual seeds; black line: mean across seeds). \textbf{b)} Learned operations. Each of the six (columns) true shift operations (top row) shift their input entries by a consistent amount. This is shown by plotting in each matrix row $i$ the outcome of applying the shift to unit vector $\ve^{(i)}$ as input. The six (columns) learned modules (bottom row) perform the same operation. \textbf{c)} Learned transitions. The true tasks incorporate a specific structure to operation sequences, shown by history-dependent transition matrices (top row). Each matrix row $i$ indicates the probability of the next module activation given that the previously selected module was module $i$ (first column), that the previous two selected modules were module $i$ (second column), previous three selected modules were module $i$ (third column), \textit{et cetera}. The learned transitions (bottom row) reproduce the true transition pattern. \textbf{d)} Example test task. After training, the model infers a solution that accurately produces (second row) the desired output for a held-out task (top row). This test task’s sequence of shift operations (third row; each matrix row represents one shift operation and red dots indicate the true underlying operation sequence) is accurately recovered by the model’s module posterior (bottom row; matrix rows are module selection probabilities so that each column plots $p(\rz_t|\rvy_{1:t})$ with red dots showing the maximum a posteriori sequence $\text{argmax}_{\rz_t}p(\rz_t|\rvy_{1:T})$. \textbf{e)} Sparse feedback example. The model still infers an accurate test task solution when feedback is provided in a small minority of timesteps (marked in yellow in the third row). \textbf{f)} Extended task example. Even on a test task that is four times longer than the training tasks, the model infers the correct solution, despite sparse feedback.}
\label{fig:rule-learning}
\end{figure}

Indeed, we find that our model successfully learns the shift operations and their sequence statistics. As the total negative log marginal likelihood over training decreases, the accuracy of the module RNNs and the gating RNN rises to plateau at 1 (Figure \ref{fig:rule-learning}a). This means that when we provide each module RNN after training with a set of one-hot $\rvy_{t-1}$ and all-zero $\rvx_t$ probe vectors, they produce outputs $\rvy_t$ with shifted entries exactly like the ground truth shift operations (Figure \ref{fig:rule-learning}b; modules are post-hoc reordered to match the order of shift operations). Moreover, when we provide the gating RNN after training with a history of module selections, the output probability distribution across modules reflects the ground truth regularities (Figure \ref{fig:rule-learning}c). The learned transition matrices highlight that after one or two repetitions of a module, that same module should be selected. After that, the gating network switches to any other module - but only for $S_0$ and $S_1$, which are each repeated 3 times. In the next step, $S_2$ and $S_3$ switch, and in the next step, $S_4$ and $S_5$ switch. Importantly, these immediate changes in the transition matrix depending on the history of module selection indicate that the gating RNN has learned the underlying strongly non-Markovian statistics; a HMM would not be able to capture those.

\subsection{One-shot task acquisition} \label{sec:inference}
With this learned generative model available, we can now infer solutions to new test tasks. We plot the posterior module selection given the data so far $p(\rz_t|\rvy_{1:t})$ at each timestep (Figure \ref{fig:rule-learning}d, bottom row, heatmap), and the model output (Figure \ref{fig:rule-learning}d, second row) for the maximum a posteriori sequence of modules $\text{argmax}_{\rz_t}p(\rz_t|\rvy_{1:T})$ (Figure \ref{fig:rule-learning}d, bottom row, red dots). These outputs show that the model reaches the correct solution, without parameter updates, based on a single data episode of the test task. But an even stronger demonstration of the power of probabilistic inference can be made under sparse feedback conditions (Figure \ref{fig:rule-learning}e). Sparse feedback means that $\rvy_t$ is available only at intermittent timesteps (Figure \ref{fig:rule-learning}e, third row, yellow timesteps). At timesteps without feedback, we cannot rely on the target likelihood to infer the appropriate module. Instead, we need to track different hypotheses on all the possible sequences of module selections until new feedback arrives. Here the gating RNN proves particularly effective: it constrains which module sequences are possible in the absence of feedback. This is reflected by the posterior $p(\rz_t|\rvy_{1:t})$ in episodes with sparse feedback (Figure \ref{fig:rule-learning}e, bottom row, heatmap). On a timestep with feedback, the posterior collapses for the subsequent timesteps without feedback - but only until the module has been repeated for the learned number of timesteps. After that, the posterior turns uniform across modules. This change is a product of the learned dynamics in the gating RNN (Figure \ref{fig:rule-learning}c) because it is not signalled by $\rvx_t$ or $\rvy_t$. As soon as feedback returns, the posterior continues only from the hypothesis confirmed by this feedback. As a result, $\text{argmax}_{\rz_t}p(\rz_t|\rvy_{1:T})$ traces the true module sequence at the end of the episode (Figure \ref{fig:rule-learning}e, bottom row, red dots). The model thus overcomes feedback sparsity through constrained hypothesis testing. This remains effective far beyond what the model was exposed to during training: the model can solve test tasks that are four times as long as any of the training tasks (Figure \ref{fig:rule-learning}f).

\begin{figure}[h]
\begin{center}
\includegraphics[scale=0.75]{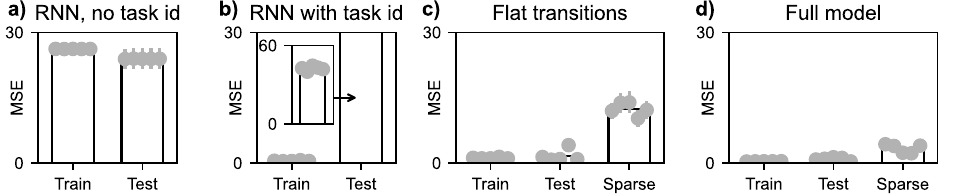}
\end{center}
\caption{Control models. \textbf{a)} RNN control model. An RNN trained to perform the tasks with the same input as our model cannot learn either train or test tasks (grey dots: individual seeds, error bars s.e.m. across tasks; black bars: mean across seeds). \textbf{b)} RNN with task identity input. When the task identity is added to the inputs, the RNN performs well on the training tasks but cannot solve new test tasks without additional training. \textbf{c)} Task inference without gating network. As our model learns a generative model across tasks, it performs well on the training tasks and infers solutions to held-out test tasks from a single episode. However, without a gating network it performs poorly on tasks with sparse feedback. \textbf{d)} Full model. The full model learns training tasks and infers test tasks even if feedback is sparse.}
\label{fig:inference}
\end{figure}

This capacity to infer a new test task from a single episode, even under sparse feedback, is specific to our architecture. We will compare it to three control models to support this claim. First, when we replace our model by a standard RNN that receives the same inputs, the RNN cannot solve the training or test tasks (Figure \ref{fig:inference}a). This is unsurprising as the network can’t know what to do across tasks without task identity input. We therefore also train a standard RNN that receives additional task identity input (Figure \ref{fig:inference}b). This control model does learn to perform the training tasks, but performs poorly on the held-out test tasks, as the task identity of the test tasks will not have been present during training. It can be retrained on the test tasks, but this will require additional weight updates through gradient descent (and even more for sparse feedback). If instead we turn to our architecture, but replace the gating RNN by a uniform transition matrix, we are able to learn the training tasks and infer the test tasks (Figure \ref{fig:inference}c). However, when feedback is sparse, the model suffers from the lack of constraints on the transition structure and fails to do task inference. Only the full architecture successfully learns the training tasks and infers the test tasks even under sparse feedback (Figure \ref{fig:inference}d).

\subsection{Learning and inferring motor skills} \label{sec:motor-learning}
\begin{figure}[h]
\begin{center}
\includegraphics[scale=0.75]{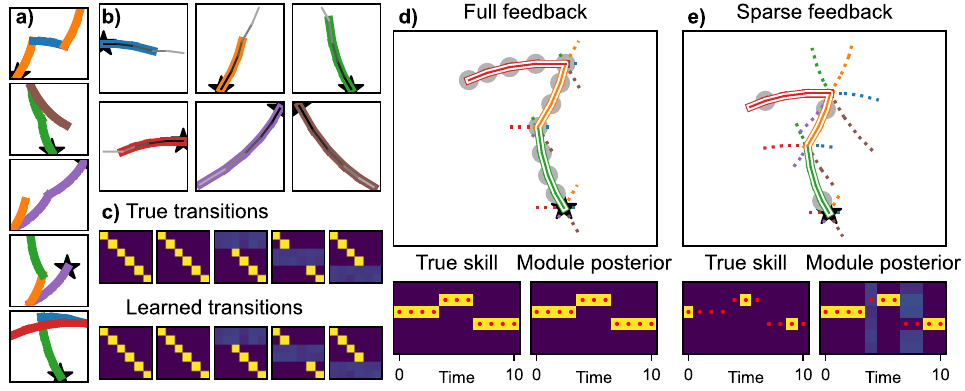}
\end{center}
\caption{Motor learning. \textbf{a)} Example motor tasks. Each task consists of a sequence of three motor skills, shown as chunks of different colours, starting from the star. \textbf{b)} Learned skills. After training, each module (output of one module in thin grey lines from dark to light in each subplot) learns to perform a skill (thick coloured lines; one true skill in each subplot). \textbf{c)} Learned transitions. The history-dependent transition matrices, analogous to Figure \ref{fig:motor-learning}c, show that the gating RNN learns to switch between skills depending on their true duration.  \textbf{d)} Example test task. The model (thin solid line coloured by module, with white borders, and pre-feedback hypotheses in dotted lines) infers the new task trajectory (thick solid line, starting from star) from feedback at each step (grey circles) by selecting learned modules (bottom right, as in Figure \ref{fig:motor-learning}d) that execute the true skills (bottom left). \textbf{e)} Sparse feedback example. When feedback is sparse (at grey circle locations, and timesteps marked in yellow at bottom left) the model tests module hypotheses (dotted lines)  that branch out at skill switch points until feedback confirms the current skill. This is reflected by the posterior $p(\rz_t|\rvy_{1:t})$ collapsing to a single module at feedback timesteps (bottom right, heatmap) and an accurate maximum a posteriori sequence $\text{argmax}_{\rz_t}p(\rz_t|\rvy_{1:T})$ (bottom right, red dots).}
\label{fig:motor-learning}
\end{figure}

Although usually considered separately from rule learning, motor learning is another domain where agents can greatly benefit from compositionality, by combining simple skills into complex behaviours \citep{ijspeertDynamicalMovementPrimitives2013a, bergSARGeneralizationPhysiological2023, merelNeuralProbabilisticMotor2019}. Our framework naturally accommodates both. In a motor task where composite paths are generated from reusable chunks, the model isolates the chunks and infers how to recombine them into new paths. We define the skills in this task as sequences of translations, each step with an increased or decreased angle compared to the previous to create curvature, and an increased or decreased magnitude to create acceleration or deceleration. Importantly, the skills have different durations: 3 steps for skill 0 and 1, 4 steps for skill 2 and 3, and 5 steps for skill 4 and 5. Tasks are then generated by concatenating three skills into a combined trajectory (Figure \ref{fig:motor-learning}a). The module RNNs should therefore each learn to output the sequence of translations for a skill, and the gating RNN must learn when to switch between skills.

Again, the model learns accurate modules and their transition statistics from the training tasks. We plot the learned skills by running each module RNN in isolation for multiple timesteps and find that the learned translations closely match the true skills (Figure \ref{fig:motor-learning}b). The module transitions learned by the gating RNN reflect the durations of the skills (Figure \ref{fig:motor-learning}c). These results illustrate how the same principles for learning a generative model can be applied to rule and motor learning tasks. Nevertheless, there are also important differences between the two. First, the motor task here does not require input $\rvx_t$, as the output is just a sequence of translations independent of inputs. Second, in the motor task each module needs to track progress within the skill, because that matters for its output. To accommodate for these differences, we make two practical changes to the model for motor learning: we remove the input $\rvx_t$ and we reset the module hidden state $\vm_t$ after a module switch. Additionally, we increase each module’s expressivity through module-specific $\mW_M^z$ and improve the efficiency of the particle filter transition proposals by sampling them from $p(\rz_t|\rz_{t-1})p(\rvy_t|\rvz_t)$ instead of $p(\rz_t|\rz_{t-1})$ during training (Appendix \ref{app:inference}).

This learned generative model of motor tasks allows us to rapidly acquire new test tasks. We plot the true skill sequence with the module posterior $p(\rz_t|\rvy_{1:t})$ as well as the maximum a posteriori sequence $\text{argmax}_{\rz_t}p(\rz_t|\rvy_{1:T})$ to demonstrate successful task inference (Figure \ref{fig:motor-learning}d, bottom). The motor task has the advantage that the model outputs are easily visualised: we show the true task path as a thick line coloured by skill, the maximum a posteriori sequence as a solid line coloured by module (which overlaps exactly with the true path, so we give it a white border), and the pre-feedback hypotheses $p(\rz_t|\rz_{t-1})$ as dotted lines coloured by module (Figure \ref{fig:motor-learning}d, top). These hypotheses are most illustrative in the sparse-feedback case (Figure \ref{fig:motor-learning}e). As before, we find that the posterior $p(\rz_t|\rvy_{1:t})$ in the absence of feedback develops through time respecting the learned module durations. But now we can also plot the content of the parallel hypotheses during these periods. As indicated by the dotted lines in Figure \ref{fig:motor-learning}e, once a module has been confirmed by feedback (grey circle), this module is continued until completion, after which different hypotheses branch out to each track the possibility of the next potential skill. As soon as feedback returns, only the confirmed hypothesis continues, and all the other branches are cut short.

\section{Discussion}
Rapidly solving new tasks relies on thinking as well as learning, but meta-learning research has often focused on the latter. Here, we propose a framework for the former. We take a compositional approach by first learning the common components and the statistics of their combinations across training tasks. We then solve new tasks from minimal experience by thinking, or more specifically, compositional inference: reasoning about potential component configurations and testing these hypotheses against the current observations. Our results provide three current contributions. First, we formalise compositional meta-learning as inference in a learned probabilistic generative model with recurrent module and gating functions. Second, we define a neural network architecture for this model and specify a procedure for training its parameters and running inference. Third, we demonstrate how our architecture solves new tasks from single examples in rule learning and motor learning domains, even when feedback is sparse, without any parameter updates.

This is in stark contrast to traditional meta-learning approaches that explicitly focus on weight updates. \citet{finnModelAgnosticMetaLearningFast2017, liLearningGeneralizeMetaLearning2018, nicholFirstOrderMetaLearningAlgorithms2018} adjust the training loop to learn parameters that require minimal gradient-based adaptation on new tasks. These methods are model-agnostic, but there are good reasons to adopt modular architectures \citep{pfeifferModularDeepLearning2024}, such as resource efficiency \citep{shazeerOutrageouslyLargeNeural2017, fedusSwitchTransformersScaling2022} and compositional generalisation \citep{changAutomaticallyComposingRepresentation2018}. In fact, modular computation emerges spontaneously in multi-task settings \citep{yangTaskRepresentationsNeural2019, driscollFlexibleMultitaskComputation2024} (although optimal modularity may not \citep{csordasAreNeuralNets2021, mittalModularArchitectureEnough2022}). Indeed, modular approaches display impressive generalisation \citep{goyalRecurrentIndependentMechanisms2020, andreasNeuralModuleNetworks2017} and meta-learning capacity \citep{martonEfficientRobustMultitask2022, dunckerOrganizingRecurrentNetwork2020, rosenbaumRoutingNetworksAdaptive2017, pontiCombiningModularSkills2022, chitnisLearningQuicklyPlan2019}. But this type of meta-learning still requires parameter updates on test tasks. The approach in \citet{aletModularMetalearning2019} doesn’t, making it most similar in spirit to ours. It fixes module parameters after training, then searches the module configuration of test tasks through simulated annealing. We effectively replace this search by probabilistic inference on learned structure, greatly improving sample efficiency. Solving new tasks without parameter updates is also related to in-activity \citep{hochreiterLearningLearnUsing2001}, in-memory \citep{santoroMetaLearningMemoryAugmentedNeural2016}, and in-context \citep{brownLanguageModelsAre2020, oswaldTransformersLearnInContext2023} learning, but those lack modularity and probabilistic reasoning - both key for compositionality in our framework. Finally, there are related brain-inspired models that exhibit rapid non-synaptic task acquisition: \citet{healdContextualInferenceUnderlies2021} though probabilistic inference, but without expressive network components and \citet{wangPrefrontalCortexMetareinforcement2018} by meta-learning the (approximate) inference algorithm itself, which only works on simple tasks.

The results reported here serve as a proof-of-principle, so there are many ways to expand on them. Importantly, the number of modules is currently predefined and fixed. As a promising direction for future work, this could be addressed by applying the model in a (class-incremental \citep{venThreeScenariosContinual2019}) continual learning setting \citep{parisiContinualLifelongLearning2019, ostapenkoContinualLearningLocal2021}. In such a setting the model would dynamically add new modules if inference using the existing modules fails. Solving new tasks through inference is particularly advantageous in continual learning because it circumvents catastrophic forgetting \citep{mccloskeyCatastrophicInterferenceConnectionist1989}, as no parameter updates are needed. Moreover, continual learning naturally affords training task curricula \citep{elioEffectsInformationOrder1984}. The current simultaneous learning of modules and gating risks training instability and local minima due to a ‘chicken-and-egg’ problem \citep{rosenbaumRoutingNetworksChallenges2019}. The gating of modules is hard to learn if their functionality hasn’t developed; the modules are hard to learn if the gating signal is inconsistent \citep{aletModularMetalearning2019}. Curriculum learning likely improves this situation \citep{changAutomaticallyComposingRepresentation2018, leeWhyAnimalsNeed2024}. Finally, we note that our architecture generalises to different gating and module components. Replacing the gating RNN by a modern transformer would allow for learning complex task grammars not unlike the one governing natural languages \citep{vaswaniAttentionAllYou2017}. Together, our results and suggestions chart a path towards the rapid composition of new solutions from learned elements across domains.



\subsubsection*{Code availability}
Code for models, training, and analysis is available for download \href{https://drive.google.com/file/d/1lOZprlKTtQEpzVtjqsC0_uAL36nGXdDt}{here}. This includes scripts to generate every panel in the figures in this paper, as well as the trained weights of all analysed models, to directly reproduce all results presented here. 

\bibliography{compositional_metalearning}
\bibliographystyle{compositional_metalearning}

\appendix
\section{Appendix}
\subsection{Model and training} \label{app:model}
We implement all RNNs, meaning the gating network and each of the module networks, as simple Elman RNNs, each with a $32$-dimensional single-layer hidden state. The control RNN in Figure \ref{fig:inference}a,b is a GRU \citep{choPropertiesNeuralMachine2014} with $7*32=224$ hidden units to match the overall number in our model (but with many more weights, as there are no independent modules). The initial hidden state of these networks is a learned parameter vector. At each timestep, a linear projection from the gating hidden state parameterises a categorical probability distribution by assigning log probabilities to each category (Equation \ref{eq:2}). We sample the selected module from these probabilities through the gumbel-softmax reparameterisation trick \citep{jangCategoricalReparameterizationGumbelSoftmax2017}. That yields a soft (i.e. approximately but not strictly one-hot) activation vector across modules, which allows for gradient flow through the sampling step during training. After training, during inference on test tasks, we replace the activation vector by its hard (i.e. strictly one-hot in the argmax dimension) equivalent. To obtain the timestep’s module hidden state, we sum the hidden state across all modules, weighted by the activation vector. A linear projection from the resulting module hidden state provides the output at the timestep (Equation \ref{eq:4}).

Training the gating and module parameters simultaneously is prone to local minima and instability. To mitigate this, we found the initialisation of network weights to be important. Small initial weights, determined by weight factor $w_{init}=0.01$ ($w_{init}=0.001$ for motor learning), decrease variability across seeds. Specifically, we initialise linear layer weights as Xavier uniform with gain $w_{init}$ and initialise linear layer bias (if present) as $0$. For recurrent networks, we initialise input weights as Xavier uniform with gain $w_{init}$ and bias as $0$; recurrent weights as orthogonal, then multiplied by $w_{init}$; and recurrent bias as $0$. We then train our model’s parameters by gradient descent on the negative marginal log likelihood (Equation \ref{eq:8}) using the ADAM optimiser \citep{kingmaAdamMethodStochastic2017} with learning rate $0.0003$ ($0.0001$ for motor learning) for $20$k iterations. We obtain this loss through particle filtering (Equations \ref{eq:5}-\ref{eq:8}; also see Appendix \ref{app:inference}) with $250$ particles. We use a batch size of $512$, gradient clipping of $1.0$, without regularisation or learning rate schedulers. 

\subsection{Probabilistic inference} \label{app:inference}
We use particle filtering to implement probabilistic inference in our learned generative model. Each particle carries the module hidden state $\vm_t$ and the gating hidden state $\vg_t$, as well as the selected module $\rz_t$. Tracking the RNN hidden states matters, because these contain long-range history dependencies; particles with the same $\rz_t$ may have different $\vm_t$ and $\vg_t$ because their module selection history differs. Additionally, we track each particle’s ancestor. After updating the particle parameters following Equations \ref{eq:1}-\ref{eq:4}, the ancestor indexes which particle the current particle was resampled from in Equation \ref{eq:6}. That allows us to trace back the sequence of module activations with the highest likelihood at the end of the episode \citep{doucForwardFilteringBackward2009}.

There are two technical details that majorly improve the efficiency of the particle resampling procedure described in Equations \ref{eq:1}-\ref{eq:6}. The first is that we use stratified (also called balanced or systematic) resampling \citep{kitagawaMonteCarloFilter1996, doucetTutorialParticleFiltering2009} to implement Equation \ref{eq:6}. Equation \ref{eq:6} defines a multinomial distribution that we repeatedly sample from to update the particle system with respect to the likelihood of the current observations. But rather than sampling from that distribution directly, we sample $\ru \sim \text{Uniform}(0, 1/K)$, and then follow a deterministic procedure for obtaining particles for the next step. We define a set of boundaries $b_k$ as the cumulative sum of the particle probabilities in Equation \ref{eq:6}: $b_k=\sum_{i=1}^k l_t^{(k)} / \sum_{j=1}^K l_t^{(j)}$ and $b_0=0$. Then we select resampled particles by taking steps $b \in \{u + i/K\}_{i=0...(K-1)}$ and returning the particles for which $b_{k-1} \leq b < b_k$. This stratified sampling reduces degeneracy, where many particles carry identical information.

The second improvement to particle resampling efficiency only applies when training the model. That is because this method, referred to as guided particle filtering \citep{carpenterImprovedParticleFilter1999}, relies on observations in the future. During inference, we assume our model doesn’t have access to these future observations, so we stick to Equations \ref{eq:1}-\ref{eq:5} (known as bootstrap particle filtering \citep{doucetIntroductionSequentialMonte2001}). In guided particle filtering, we sample particle module selection $\rz_t \sim p(\rz_t|\rz_{t-1})p(\rvy_t|\rz_t)$ instead of $\rz_t \sim p(\rz_t|\rz_{t-1})$ as implemented by Equation \ref{eq:2}. That means Equations \ref{eq:1}-\ref{eq:5} are replaced by 
\begin{equation}
\vg_t=G_\vtheta(\rvx_t,\vg_{t-1},\rz_{t-1})
\end{equation}
\begin{equation}
\vm^{z_t}_t=M^{z_t}_\vphi(\rvx_t,\vm_{t-1})
\end{equation}
\begin{equation}
f_{z_t}=L_{\text{Cat}}(\rz_t;\mW_G \vg_t) L_{\text{MVN}}(\rvy_t;\mW_M \vm^{z_t}_t, \sigma \mI)
\end{equation}
\begin{equation}
\rz_t \sim \text{Cat}(f_z / \sum_{i=1}^N f_i)
\end{equation}
\begin{equation}
l_t^{(k)}=\sum_{i=1}^N f^{(k)}_i
\end{equation}
where $L_{\text{dist}}(x;\lambda)$ denotes the likelihood of $x$ under distribution $dist$ with parameters $\lambda$. We use guided particle filtering only for the motor learning task in the current work, but it is task-agnostic.
\end{document}